# Improving Anomaly Detection in Industrial Time Series: The Role of Segmentation and Heterogeneous Ensemble

E. Mastriani, A. Costa, F. Incardona, K. Munari, *S. Spinello*

*Abstract*—Concerning machine learning, segmentation models can identify state changes within time series, facilitating the detection of transitions between normal and anomalous conditions. Specific techniques such as Change Point Detection (CPD), particularly algorithms like ChangeFinder, have been successfully applied to segment time series and improve anomaly detection by reducing temporal uncertainty, especially in multivariate environments.

In this work, we explored how the integration of segmentation techniques, combined with a heterogeneous ensemble, can enhance anomaly detection in an industrial production context. The results show that applying segmentation as a pre-processing step before selecting heterogeneous ensemble algorithms provided a significant advantage in our case study, improving the AUC-ROC metric from 0.8599 (achieved with a PCA and LSTM ensemble) to 0.9760 (achieved with Random Forest and XGBoost). This improvement is imputable to the ability of segmentation to reduce temporal ambiguity and facilitate the learning process of supervised algorithms.

In our future work, we intend to assess the benefit of introducing weighted features derived from the study of change points, combined with segmentation and the use of heterogeneous ensembles, to further optimize model performance in early anomaly detection.

## I. Introduction

In recent years, anomaly detection in time series has become a critical issue in the industrial context. Timely identification of anomalous behaviors can prevent critical failures, reduce downtime, and considerably improve overall operational efficiency. However, accurate anomaly detection is made challenging by the multivariate nature of sensor data and the inherent uncertainty in the temporal labels provided by domain experts. Often, it is not possible to collect precise information about the exact failure dates, but only indicative time intervals alternating between normal and anomalous states. Further, the declared failure periods are often a low percentage of the available period, making anomaly identification even more complex.

To address these challenges, various segmentation techniques have been proposed to reduce temporal uncertainty and improve the effectiveness of detection models' [1]. Among these, CPD methods [2], such as ChangeFinder [3], have proven especially effective in identifying significant transitions between different operational states, providing a useful pre-processing step for supervised machine learning models. The adoption of segmentation models can improve prediction accuracy compared to traditional anomaly detection approaches. Specifically, ChangeFinder is an online and unsupervised algorithm for anomaly detection and change point detection in time series data. It is notably efficient in identifying sudden changes in statistical properties, such as shifts in mean or variance.

In the realm of predictive maintenance, recent studies have highlighted that time series segmentation can provide crucial information for detecting transitions between normal and anomalous states [4].

Meanwhile, the use of heterogeneous ensemble learning, which combines models with complementary characteristics, has shown promising results in enhancing the robustness and accuracy of anomaly detection. However, the effectiveness of these models significantly depends on the quality of the features used, and the ability to isolate relevant states correctly changes [5].

### A. Study Objectives

The primary purpose of this work is to improve anomaly detection in the industrial context through a combined approach that integrates segmentation techniques and supervised machine learning models.

### B. Significance of the Work

This research offers new perspectives for enhancing anomaly detection techniques in industrial time series, contributing to the development of more accurate and robust methodologies for predictive maintenance. The study aims to highlight the crucial role of segmentation and heterogeneous ensemble learning in improving anomaly detection predictions, to reduce false positives and false negatives in industrial settings. In addition, future studies will focus on introducing weighted features derived from change point analysis, to further optimize model performance and pave the way for innovative solutions in predictive machinery management.

### C. Context and Dataset

The equipment studied in this work is a steam turbine connected to an electric generator. The apparatus is placed in an industrial plant, which is fully digitalized. The steam turbine converts the pressure drop from high-pressure steam (HP) to medium-pressure steam (MP) into electrical energy, effectively recovering energy. In case of turbine unavailability, the steam can be diverted through a dedicated valve (within the turbine package) that reduces the steam pressure from HP to MP. The electricity production of this turbine is directly linked to the steam demand of the plant's utilities and,

E. Mastriani (phone: +39 095 7332281; fax: +39 095 330592; e-mail: emilio.mastriani@ inaf.it), A. Costa (alessandro.costa@inaf.it), F. Incardona (filippo.incardona@inaf.it), K. Munari (kevin.munari@inaf.it), and S. Spinello

Authors are with INAF, Osservatorio Astrofisico di Catania, Via S. Sofia 78, I-95123 Catania, Italy (Sebastiano.spinello@inaf.it).

therefore, to the production level of the refining center. With 70 variables (features), the dataset in question contains 1,124,820 data points. The training data covers the period from July 9, 2022, to August 3, 2023, while the test data covers the period from September 1, 2023, to November 22, 2024. The confirmed anomaly range spans from August 11, 2024, to August 17, 2024. This 7-day time interval represents approximately 1.56% of the total test dataset duration, which covers 448 days. The data frame object in this study is accompanied by a Normal Operating Condition (NoC) file. This file identifies the periods during which the turbine was assessed working under normal conditions by industry experts. The NoC file was used for two primary purposes: a. identifying machine operating and idle periods and treating them as a sequence of sets along the temporal axis, to quantify the effectiveness of the segmentation used; b. using the operating state of the compressor (normal/anomalous) as the target feature during the training of hybrid models.

The dataset is naturally imbalanced, with a prevalence of "normal" data compared to anomalous events. Therefore, before assessing the contribution of the segmentation technique, models capable of handling imbalanced situations were prioritized.

## II. METHODOLOGY

### A. Data Preparation Procedure for Predictive Maintenance

The process starts with a complex dataset containing over one million observations and 70 features collected from industrial sensors, which requires significant cleaning and transformation before it can be used for machine learning models.

The first phase involves data cleaning, where two entirely null features are removed, and anomalous negative values are replaced with the median, chosen for its robustness to outliers. Missing values are handled similarly, ensuring the preservation of the original feature distribution.

The next step is data normalization. Through a detailed analysis of skewness and kurtosis, the features are divided into three categories: those with near-symmetric distribution (requiring no intervention), moderately skewed features (transformed using Yeo-Johnson [6] to improve distribution), and highly skewed features or those with diverse outliers (where more aggressive techniques such as Winsorization are applied). This differentiated approach allows the data to be better suited to the assumptions of machine learning algorithms without excessively altering its nature.

Feature selection is performed in multiple steps. First, low-variance, non-informative features are eliminated. Then, through ANOVA and Mutual Information [7], the most relevant features correlated with the anomalous operation of the machine (target "Normal=False") are identified.

Finally, collinearity between features is reduced by retaining only the most significant ones and eliminating redundancies that could impair the model.

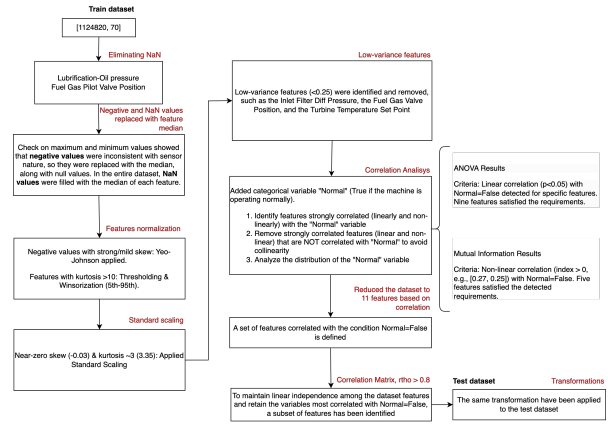

*Figure 1 Data Cleaning and Preparation Workflow. Flowchart illustrating the data preparation process for predictive maintenance. Starting from a large industrial sensor dataset, the procedure includes data cleaning, differentiated normalization based on distribution analysis, and multi-step feature selection to ensure data quality and model readiness.*

The outcome is an optimized dataset, with relevant features, normalized distributions, and balanced classes, ready to train classification. The entire process is replicable on new data, ensuring consistency between the development and production phases. The complete process is described by the flowchart shown in Fig. 1.

### B. Procedure for Developing non-Hybrid Classification Models Without Segmentation

The core of this procedure lies in transforming raw data from industrial sensors into an intelligent system capable of predicting machine failures. The approach combines advanced temporal processing techniques with machine learning models, following a structured process that begins with data organization and concludes with the prediction of critical states.

*1) Temporal Data Processing*
The first critical step involves the temporal reprocessing of data. Since sensors continuously generate information, it is essential to organize these streams into meaningful temporal windows. By employing the sliding window technique [8], the data are divided into consecutive blocks (30-minute intervals), each serving as input to predict the operational state in the immediate future. This method ensures that the model learns not only from absolute values but, more importantly, from the temporal dynamics preceding a failure. For each window, a label distinguishing between normal and anomalous operation is computed, thus creating a supervised dataset.

*2) Feature Extraction*
The second phase focuses on extracting targeted features from each temporal window. Besides traditional descriptive statistics, more sophisticated indicators, such as trends, derivatives, and the frequency of critical threshold exceedance, are calculated. These features capture both the instantaneous state of the machinery and its evolution over time, providing the model with early warning signals of potential anomalies.

*3) Temporal Separation Between Training and Testing Sets*

The innate temporal separation between training and testing datasets ensures a realistic assessment of model performance, accurately simulating the operational conditions under which the model will function. This approach eliminates the need for additional time-based splitting of the dataset.

*4) Considered Classification Models*

Hybrid classification models were evaluated to assess their performance metrics. This analysis includes traditional algorithms such as Support Vector Machine (SVM), Random Forest, and Gradient Boosting, combined with advanced feature extraction and selection techniques.

## C. Procedure for Developing Non-hybrid Classification Models with Segmentation

ChangeFinder is an unsupervised algorithm used for anomaly detection and change point identification in time series. ChangeFinder is especially well-suited for detecting rare and subtle changes in highly imbalanced time series data, as in our case. Its lightweight and fully online structure enables it to operate efficiently in real-time environments, making it ideal for monitoring applications where faults are infrequent but critical. The simplicity and low computational overhead render it a practical and effective solution for early fault detection in streaming sensor data. One of its key strengths is the ability to detect anomalies in real time with no labeled training data, making it especially valuable in scenarios where data is continuous and unlabeled. A crucial aspect of the algorithm is the tuning of parameters, which plays a key role in determining the model's effectiveness in detecting changes and anomalies. Specifically, three parameters are fundamental: r (learning rate), which controls how quickly the model updates; order (ARIMA), which sets AR, I, and MA terms for time series modeling; and smooth (window size), which balances sensitivity and responsiveness by adjusting the smoothing window.

Optimizing these parameters is crucial for improving ChangeFinder's ability to detect change points and anomalies, reducing false positives, and improving the model's reliability in practical applications. To identify the optimal parameter set, we implemented a pipeline that runs in parallel to search for the optimal values of the parameters r, order, and smooth through parallel grid search. ChangeFinder algorithm is used to detect change points and assign a score that measures the quality of anomaly detection concerning state changes in the "Normal" variable (from True to False). The optimization function, which explores different parameter combinations, calculates the F1 score and CS (Change Score) for each to determine the parameter combination that maximizes the performance of the anomaly detection model. In this context, the CS and the F1 scores were compared, resulting in two different data segmentations, as shown in Fig. 2 and Fig. 3. In the graphs, the blue line shows the trend of the feature values over time, while the orange dashed lines show the points where significant changes in the operational state were detected, and the sections in pale yellow highlight the periods of confirmed anomalies.

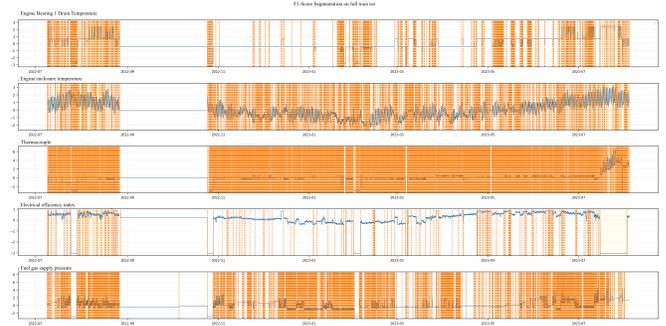

*Figure 2 Segmentation achieved with optimal parameters based on F1-score: 'r': 0.05, 'order': 1, 'smooth': 5, 'threshold': 1.8*

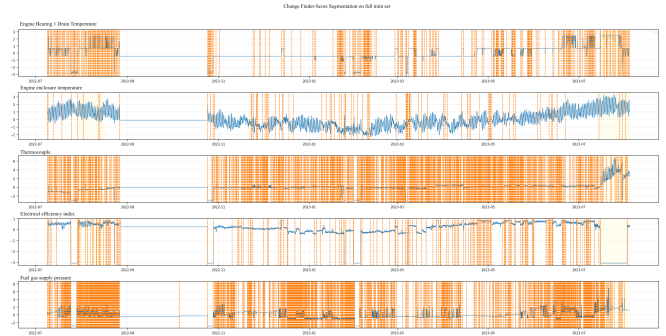

*Figure 3 Segmentation achieved with optimal parameters based on Change Finder score: 'r': 0.1, 'order': 1, 'smooth': 10, 'threshold': 1.5*

Finally, models in both hybrid and non-hybrid forms were executed to assess their metrics: Random Forest[9], Isolation Forest[10], XGBoost[11], LSTM[12], PCA[13], KMeans[14], One-Class SVM[15], and Auto Encoder[16].

## III. RESULTS

### A. Preventive Maintenance Strategy Without Segmentation

The data in Table 1 shows the main metrics gained from executing non-hybrid models performed without segmentation. In particular, among the tested non-hybrid models, two approaches stood out for their complementary characteristics.

The PCA-based model exhibited remarkable performance, achieving the highest AUC-ROC score among all non-hybrid models, with a value of 0.8198. This indicator, particularly significant in imbalanced classification contexts, suggests that the algorithm can distinguish between normal states and anomalies with strong discriminatory power. Although the absolute values of precision (0.17) and recall (0.16) for class 1 (anomalies) may seem modest, the reasonable F1-score (0.17) suggests an acceptable balance between these two metrics. These results imply that PCA, despite its limitations in capturing all anomalies, remains a valid approach for detecting certain anomalous states.

Conversely, the LSTM model shows different characteristics. With a slightly lower F1-score (0.13) and a lower AUC-ROC (0.4733), its performance appears less impressive in anomaly detection.

However, it is important to note that this model excels, particularly in recognizing class 0 (normal states), where it achieves very high precision and recall. This characteristic still makes it one of the best non-hybrid models, especially in scenarios where correctly identifying normal states is a priority. Therefore, we considered the hybrid approach that combines PCA with LSTM. This strategy proved to be the most effective among all those tested, demonstrating particular strengths but also some limitations to consider.

*Table 1 Metrics of non-hybrid models without segmentation. The table presents the performance metrics (Precision, Recall, F1-score, and AUC-ROC) for various non-hybrid models without segmentation. The models are evaluated across two cases: fault (1) and non-fault (0). Key observations include high precision for models like IF, RF, and PCA in non-fault cases, while recall and F1 scores vary significantly across the different models for fault cases.*

| Model | Case (fault) | Precision | Recall | f1-score | AUC-ROC |
|---|---|---|---|---|---|
| IF | 0 | 0.99 | 0.75 | 0.85 | 0.6873 |
|  | 1 | 0.03 | 0.45 | 0.05 |  |
| RF | 0 | 0.99 | 0.91 | 0.95 | 0.65 |
|  | 1 | 0.05 | 0.26 | 0.08 |  |
| XGBoost | 0 | 0.99 | 0.07 | 0.13 | 0.3176 |
|  | 1 | 0.02 | 0.98 | 0.03 |  |
| LSTM | 0 | 0.99 | 0.98 | 0.98 | 0.4733 |
|  | 1 | 0.11 | 0.15 | 0.13 |  |
| KMeans | 0 | 0 | 0 | 0 | 0.2354 |
|  | 1 | 0.02 | 1 | 0.03 |  |
| PCA | 0 | 0.99 | 0.99 | 0.99 | 0.8198 |
|  | 0 | 0.17 | 0.16 | 0.17 |  |
| Auto Encoder | 0 | 0.99 | 0.8 | 0.89 | 0.7236 |
|  | 1 | 0.04 | 0.49 | 0.07 |  |
| One Class SVM | 0 | 0 | 0 | 0 | 0.74 |
|  | 1 | 0.02 | 1 | 0 |  |

The most significant value is the AUC-ROC of 0.8599, the highest achieved by any model in our study without segmentation. This result argues that the combination of these two techniques has a good capacity to distinguish between normal states and anomalies, surpassing both individual models and other hybrid configurations. The high AUC-ROC suggests that the model is particularly adept at correctly classifying both positive and negative examples, assigning higher scores to true anomalies compared to false alarms.

However, the analysis of metrics specific to Class 1 (anomalies) reveals some issues. With a precision of 0.08 and a recall of 0.28, the model shows a tendency to produce a significant number of false positives (low precision), and simultaneously, a non-negligible ability to capture true anomalies (relatively higher recall). This combination results in a modest F1-score (0.12) for the anomaly class, highlighting how the model still struggles to achieve an optimal balance between correctly identifying anomalies and limiting false alarms.

B. *Preventive Maintenance Strategy with Segmentation*

Table 2 shows the performance metrics differences gained from the execution of non-hybrid models performed with the help of the segmentation, according to the optimal parameters achieved with the F1 and CS scores, respectively. In particular, the comparison of the AUC-ROC metric from the two approaches, as shown in the graph in Fig. 4, suggests that models based on segmentation reached using CS scoring perform better.

*Table 2 Performance metric differentials of non-hybrid models between F1-score and Change Score-based segmentation. Each value shows the pointwise difference per model and case (0 = normal, 1 = fault); positive values favor F1-score optimization, and negatives favor Change Score.*

| Model | Case (fault) | Δ Precision | Δ Recall | Δ f1-score | Δ AUC-ROC |
|---|---|---|---|---|---|
| IF | 0 | 0.00 | -0.05 | -0.04 | -0.0627 |
|  | 1 | -0.01 | -0.01 | -0.02 |  |
| RF | 0 | -0.01 | 0.10 | 0.06 | -0.31 |
|  | 1 | -0.03 | -0.74 | -0.06 |  |
| XGBoost | 0 | 0.00 | -0.79 | -0.79 | -0.5524 |
|  | 1 | -0.05 | 0.36 | -0.09 |  |
| LSTM | 0 | 0.00 | 0.14 | 0.07 | -0.2867 |
|  | 1 | 0.07 | -0.27 | 0.06 |  |
| KMeans | 0 | 0.00 | 0.00 | 0.00 | 0.0254 |
|  | 1 | 0.00 | 0.00 | 0.00 |  |
| PCA | 0 | -0.01 | 0.21 | 0.11 | 0.0398 |
|  | 1 | 0.10 | -0.84 | 0.04 |  |
| Auto Encoder | 0 | 0.00 | -0.19 | -0.10 | -0.0864 |
|  | 1 | -0.21 | 0.22 | -0.19 |  |
| One Class SVM | 0 | 0.00 | 0.00 | 0.00 | -0.04 |
|  | 1 | 0.00 | 0.00 | -0.03 |  |

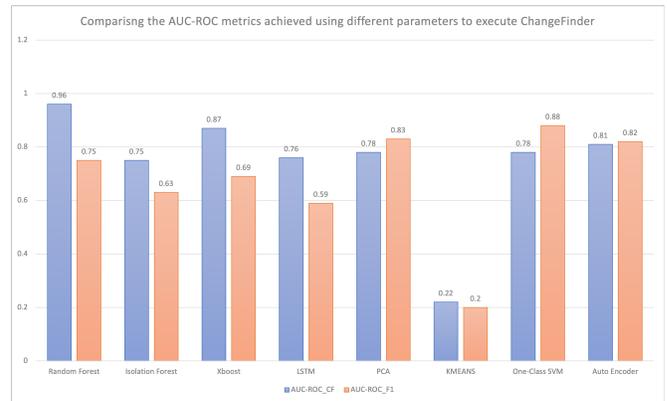

*Figure 4 The AUC-ROC metric comparison after segmentation*

A comparative analysis of the results highlights the distinctive characteristics of RF and XGBoost compared to other models tested. Both exhibit interesting predictive capabilities, albeit with slightly different approaches and outcomes.

Random Forest stands out for its ability to identify anomalies, as evidenced by its perfect recall on class 1, which reaches 1.0. This means the model successfully captured all failure situations present in the dataset. The AUC-ROC of 0.96, second only to the autoencoder, confirms the model's discriminative power in correctly separating normal states from anomalies. However, this sensitivity to anomalies comes at a cost: the precision of 0.08 emphasizes that only 8% of the positive predictions were correct, resulting in a high false alarm rate.

XGBoost shows similar performance but with a slightly better balance across different metrics. With a recall of 0.62 on class 1, it does not achieve the complete coverage of RF, but it still maintains a good anomaly detection capacity. The precision of 0.07 and the F1-score of 0.12 suggest that XGBoost better manages the trade-off between anomaly detection and limiting false positives. The

AUC-ROC of 0.87, though lower than RF's, remains one of the highest recorded.

### C. Comparison of Optimal Hybrid Solutions

The comparative analysis of the data presented in Table 3 and depicted in Fig. 5 for the two hybrid models reveals significant differences in their predictive capabilities, with each approach displaying distinct strengths relative to the different evaluation metrics.

The RF+XGBoost hybrid model emerges as the overall top performer, demonstrating marked superiority in almost all key metrics. With an exceptional AUC-ROC of 0.9760, this approach shows near-perfect discriminative ability in distinguishing between normal states and anomalies. Its performance is notably significant for class 1 (faults), where it achieves a recall of 0.69 and precision of 0.29, resulting in an F1-score of 0.41, significantly higher than the PCA+LSTM competitor.

The PCA+LSTM model, while showing a respectable AUC-ROC of 0.8599, exhibits clear limitations in anomaly detection. The recall of 0.28 for class 1 reveals that this model captures less than a third of actual anomalies, while the precision of 0.08 signals a high false positive rate. These values translate into a modest F1-score of 0.12 for fault situations.

The most significant difference between the two models is observed in their ability to balance precision and recall. While RF+XGBoost achieves a reasonable balance (precision 0.29 and recall 0.69), PCA+LSTM displays a more pronounced imbalance, with extremely low precision despite a still limited recall. Regarding class 0 (normal states), both models perform excellently, with RF+XGBoost maintaining a slight advantage in recall (0.97 vs. 0.95) and F1-score (0.98 vs. 0.97).

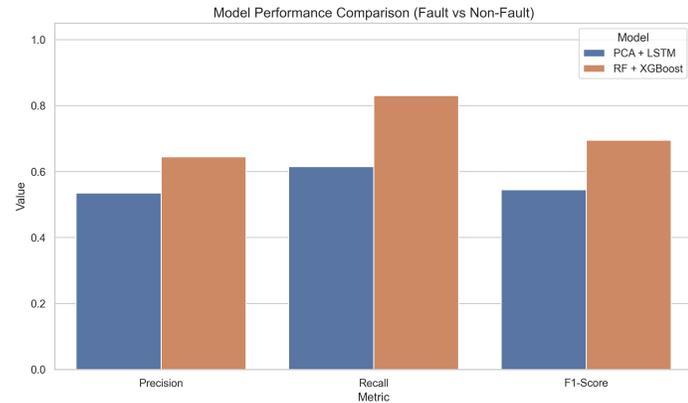

*Figure 5 Comparison of the average classification performance between two models: PCA + LSTM and RF + XGBoost. The bars represent the mean Precision, Recall, and F1-Score across fault (class 1) and non-fault (class 0) cases. The RF + XGBoost model consistently outperforms PCA + LSTM across all three metrics, particularly in detecting fault events*

*Table 3 Comparative Analysis of Hybrid Models: Segmented vs. Non-Segmented Datasets. The table compares hybrid models, showing the performance metrics (Precision, Recall, F1-score, and AUC-ROC) for both segmented datasets (highlighted with a yellow background) and non-segmented datasets (highlighted with a gray background). The models evaluated are PCA + LSTM and RF + XGBoost.*

| Model | Case (fault) | Precision | Recall | f1-score | AUC-ROC |
|---|---|---|---|---|---|
| PCA + LSTM | 0 | 0.99 | 0.95 | 0.97 | 0.8599 |
|  | 1 | 0.08 | 0.28 | 0.12 |  |
| RF + XGBoost | 0 | 1 | 0.97 | 0.98 | 0.976 |
|  | 1 | 0.29 | 0.69 | 0.41 |  |

### D. The Health Index and its Application in Predictive Maintenance

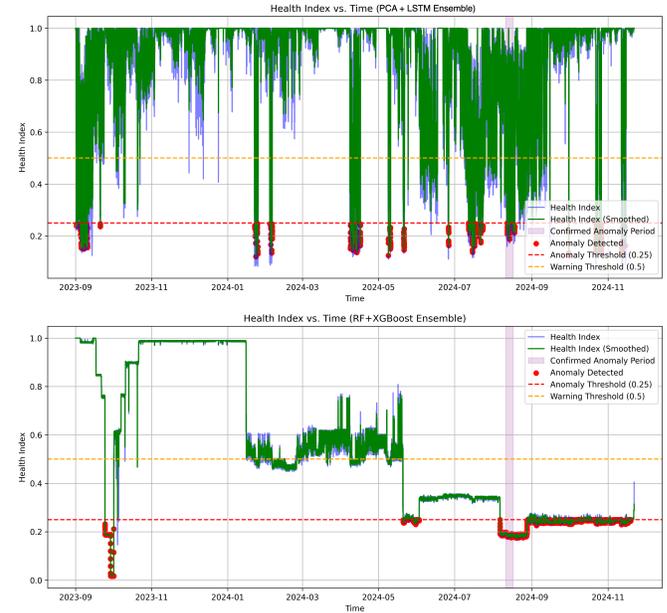

*Figure 6 Hybrid models comparison by Health Index: The Top shows PCA+LSTM (unsegmented), the bottom shows RF+XGBoost (segmented).*

The Health Index [17] is a crucial indicator for assessing the condition of an industrial asset, providing a clear and immediate measure of its operational status. This index, ranging from 0 to 1 (where 0 means an imminent failure and 1 represents an optimal condition), is calculated by inverting the anomaly probabilities collected from a hybrid model combining RF and XGBoost, as well as PCA and LSTM. In the generated graphs, the Health Index is displayed alongside two critical thresholds that are useful for interpreting its values. The first threshold, set at 0.5 and represented by a dashed orange line, serves as a preliminary warning. When the index falls below this value, it signals a deterioration in the asset's condition, suggesting the need for closer monitoring to prevent potential failures. The second threshold, set at 0.25 and marked by a dashed red line, represents a high alert level. A Health Index below this threshold implies a tangible risk of malfunction, requiring immediate intervention to avoid damage or downtime. The graphs also show a purple period corresponding to a confirmed anomaly interval between August 11 and 17, 2024, during which the asset experienced confirmed issues. The red points scattered along the curve illustrate the moments when the system automatically detected anomalies, with the Health Index dropping below 0.25. To enhance the readability of the trend, the index is smoothed using a moving average, represented by the green curve, which filters out data noise and highlights long-term trends. By comparing the two Health Index graphs presented in Fig. 6, it seems likely that the model based on the hybrid approach of RF and XGBoost, following the segmentation achieved through ChangeFinder, shows a step right at the confirmed anomaly, demonstrating greater accuracy compared to the hybrid PCA and LSTM model without segmentation.

## IV. CONCLUSIONS

This research has demonstrated how the combination of temporal segmentation techniques (ChangeFinder) with heterogeneous machine learning ensembles can significantly improve anomaly detection in complex industrial contexts. The proposed approach, combining ChangeFinder's ability to identify state changes with the predictive power of hybrid models like RF + XGBoost, resulted in a notable performance improvement, achieving an AUC-ROC of 0.9760 compared to 0.8599 with a traditional ensemble (PCA + LSTM). The results confirm that preliminary data segmentation reduces temporal uncertainty and facilitates the learning of supervised models, enabling more precise detection of transitions between normal and anomalous states. In particular, Changefinder achieves precise segmentation in balanced time series (<2% anomalies) by combining adaptive AR modeling with change-point detection, effectively isolating anomalies for accurate classification. Using it to identify change points optimized the selection of critical time windows, improving both recall (0.69 vs. 0.28) and F1-score (0.41 vs. 0.12) compared to non-segmented solutions. The introduction of the Health Index, derived from the anomaly probabilities of the hybrid model, further enhanced the interpretability of the results, providing a clear and scalable indicator of the asset's status. The operational thresholds (0.5 for preliminary alert and 0.25 for urgent intervention) proved effective in predicting failures, as evidenced by the accurate identification of the confirmed anomaly period (August 11-17, 2024). To further optimize the system, upcoming steps will include weighting features via change point analysis, exploring advanced hybrid models combining deep learning and ensembles, and adapting dynamic Health Index thresholds to the operational context. In conclusion, this work provides a solid methodological framework for predictive maintenance, demonstrating that the synergy between temporal segmentation and hybrid models can enhance the reliability of industrial monitoring systems, reducing costs and machine downtime.

## ACKNOWLEDGMENT

This work is supported by ICSC – Centro Nazionale di Ricerca in High Performance Computing, Big Data and Quantum Computing, funded by the European Union – NextGenerationEU. The authors gratefully acknowledge Alfonso Amendola for his technical and scientific assistance.

## REFERENCES

[1] C. Wang, X. Li, T. Zhou, and Z. Cai, "Unsupervised Time Series Segmentation: A Survey on Recent Advances," *Computers, Materials and Continua,* vol. 80, no. 2, pp. 2657-2673, 2024/08/15/ 2024, doi: https://doi.org/10.32604/cmc.2024.054061.

[2] S. Aminikhanghahi and D. J. Cook, "A Survey of Methods for Time Series Change Point Detection," *Knowl Inf Syst,* vol. 51, no. 2, pp. 339-367, May 2017, doi: 10.1007/s10115-016-0987-z.

[3] K. Yamanishi and J. i. Takeuchi, "A unifying framework for detecting outliers and change points from non-stationary time series data," *Proceedings of the eighth ACM SIGKDD international conference on Knowledge discovery and data mining,* 2002.

[4] D. Coelho, D. Costa, E. M. Rocha, D. Almeida, and J. P. Santos, "Predictive maintenance on sensorized stamping presses by time series segmentation, anomaly detection, and classification algorithms," *Procedia Computer Science,* vol. 200, pp. 1184-1193, 2022/01/01/ 2022, doi: https://doi.org/10.1016/j.procs.2022.01.318.

[5] Z. Wang, Y. Hong, L. Huang, M. Zheng, H. Yuan, and R. Zeng, "A comprehensive review and future research directions of ensemble learning models for predicting building energy consumption," *Energy and Buildings,* vol. 335, p. 115589, 2025/05/15/ 2025, doi: https://doi.org/10.1016/j.enbuild.2025.115589.

[6] I. K. Yeo and R. A. Johnson, "A new family of power transformations to improve normality or symmetry," *Biometrika,* vol. 87, no. 4, pp. 954-959, 2000, doi: 10.1093/biomet/87.4.954.

[7] N. Carrara and J. Ernst, "On the Estimation of Mutual Information," *Proceedings,* vol. 33, no. 1, p. 31, 2019. [Online]. Available: https://www.mdpi.com/2504-3900/33/1/31.

[8] V. Braverman, "Sliding Window Algorithms," in *Encyclopedia of Algorithms*, M.-Y. Kao Ed. New York, NY: Springer New York, 2016, pp. 2006-2011.

[9] L. Breiman, "Random Forests," *Machine Learning,* vol. 45, no. 1, pp. 5-32, 2001/10/01 2001, doi: 10.1023/A:1010933404324.

[10] F. T. Liu, K. M. Ting, and Z. H. Zhou, "Isolation Forest," in *2008 Eighth IEEE International Conference on Data Mining*, 15-19 Dec. 2008 2008, pp. 413-422, doi: 10.1109/ICDM.2008.17.

[11] T. Chen and C. Guestrin, "XGBoost: A Scalable Tree Boosting System," presented at the Proceedings of the 22nd ACM SIGKDD International Conference on Knowledge Discovery and Data Mining, San Francisco, California, USA, 2016. [Online]. Available: https://doi.org/10.1145/2939672.2939785.

[12] G. Van Houdt, C. Mosquera, and G. Nápoles, "A review on the long short-term memory model," *Artificial Intelligence Review,* vol. 53, no. 8, pp. 5929-5955, 2020/12/01 2020, doi: 10.1007/s10462-020-09838-1.

[13] M. Greenacre, P. J. F. Groenen, T. Hastie, A. I. D'Enza, A. Markos, and E. Tuzhilina, "Principal component analysis," *Nature Reviews Methods Primers,* vol. 2, no. 1, p. 100, 2022/12/22 2022, doi: 10.1038/s43586-022-00184-w.

[14] X. Jin and J. Han, "K-Means Clustering," in *Encyclopedia of Machine Learning*, C. Sammut and G. I. Webb Eds. Boston, MA: Springer US, 2010, pp. 563-564.

[15] S. Alam, S. K. Sonbhadra, S. Agarwal, and P. Nagabhushan, "One-class support vector classifiers: A survey," *Knowledge-Based Systems,* vol. 196, p. 105754, 2020/05/21/ 2020, doi: https://doi.org/10.1016/j.knosys.2020.105754.

[16] K. Berahmand, F. Daneshfar, E. S. Salehi, Y. Li, and Y. Xu, "Autoencoders and their applications in machine learning: a survey," *Artificial Intelligence Review,* vol. 57, no. 2, p. 28, 2024/02/03 2024, doi: 10.1007/s10462-023-10662-6.

[17] K. Bajarunas, M. L. Baptista, K. Goebel, and M. A. Chao, "Health index estimation through integration of general knowledge with unsupervised learning," *Reliability Engineering & System Safety,* vol. 251, p. 110352, 2024/11/01/ 2024, doi: https://doi.org/10.1016/j.ress.2024.110352.